\documentclass[runningheads]{llncs}
\usepackage[T1]{fontenc}
\usepackage{graphicx}
\usepackage{multirow}
\usepackage{xcolor}
\usepackage{amsmath, amssymb, amsfonts}
\usepackage{bm}
\usepackage{mathtools}
\usepackage{booktabs}
\usepackage{algorithm}
\usepackage{algorithmic}
\usepackage{wrapfig}
\begin{document}
\title{RS-Diffuser: Risk-Sensitive Diffusion Planning with Distributional Value Guidance}
%
%

\author{Shiqiang Gong\inst{1}\orcidID{0009-0006-9412-9949}}

\authorrunning{S. Gong}

\institute{Northwestern Polytechnical University\\
\email{gongshiqiang@mail.nwpu.edu.cn}}

%
\maketitle              
\begin{abstract}
Offline reinforcement learning enables policy learning from fixed datasets without additional environment interaction, making it appealing for safety-critical applications where online exploration is costly or unsafe. Diffusion-based decision-making methods have recently achieved strong performance in offline RL by modeling rich, multimodal trajectory distributions. However, existing diffusion planners are typically risk-neutral and therefore may overlook rare but catastrophic outcomes that are crucial in real-world deployment. In this work, we propose \textbf{RS-Diffuser}, a risk-sensitive offline diffusion planning framework that combines diffusion-based trajectory generation with distributional value critics. RS-Diffuser learns a diffusion planner over future state trajectories, a separate inverse dynamics model for action decoding, and a Monte Carlo distributional critic that estimates the full return distribution of candidate plans through quantile regression. At sampling time, we incorporate a risk-sensitive guidance signal into the denoising process, using gradients computed from tail-aware objectives such as Conditional Value at Risk to steer generation toward desired risk profiles. As a result, a single trained model can flexibly produce risk-averse, risk-neutral, or risk-seeking behaviors by changing only the inference-time risk parameter. Extensive experiments on risk-sensitive D4RL and risky robot navigation benchmarks demonstrate that RS-Diffuser achieves state-of-the-art performance, improving both overall return and worst-case robustness while reducing safety violations.

\end{abstract}

\section{Introduction}
\vspace{-0.1in}
Offline reinforcement learning (RL) seeks to learn effective decision-making policies from a fixed dataset without further interaction with the environment~\cite{prudencio2023survey}. This setting is particularly attractive in safety-critical domains such as healthcare, autonomous driving, and robotics, where online exploration is costly, unsafe, or impractical~\cite{wang2022deep}. However, offline RL remains fundamentally challenging due to distributional shift: the learned policy must act beyond the support of the behavior data, often leading to unreliable value estimates and poor generalization. In addition, many existing offline RL methods rely on deterministic or simple Gaussian policy classes, which can be insufficient for modeling the complex and multimodal behaviors commonly present in real-world datasets~\cite{zhu2023diffusion}.

Recently, diffusion models have emerged as a powerful paradigm for offline decision making by casting control as conditional generative modeling over actions or trajectories~\cite{ho2020denoising,zhu2023diffusion}. Existing approaches can be broadly divided into two families. Planning-based methods, such as Diffuser~\cite{janner2022planning}, generate trajectories through iterative denoising and perform decision making by trajectory optimization. Policy-based methods, such as Diffusion-QL~\cite{wang2022diffusion}, instead use diffusion models as expressive policy classes for direct action generation. By modeling rich multimodal distributions and long-horizon dependencies, diffusion methods have achieved strong performance across a range of offline RL benchmarks.

Despite these advances, most existing diffusion-based decision-making methods remain fundamentally risk-neutral: they optimize expected return and do not explicitly model the tail behavior of returns. This limitation is particularly problematic in safety-critical settings, where two candidate behaviors may have similar expected returns but drastically different downside risk~\cite{shen2014risk}. Risk-sensitive RL addresses this issue by optimizing risk-aware objectives rather than expectation alone~\cite{xu2024uncertainty,noorani2025risk}. Among these approaches, distributional RL provides a principled foundation by explicitly modeling the full return distribution, thereby enabling tail-aware criteria such as Value-at-Risk and Conditional Value-at-Risk (CVaR)~\cite{bellemare2017distributional,bellemare2023distributional,dabney2018distributional}. However, existing risk-sensitive offline RL methods are predominantly built on actor-critic or policy-optimization pipelines, and relatively few works have explored how to endow diffusion-based planners with explicit and controllable risk sensitivity~\cite{chen2025diffusion,miao2025uncertaintyaware,xu2025distributional}.

In this paper, we propose \textbf{RS-Diffuser}, a risk-sensitive offline diffusion planning framework that combines diffusion-based trajectory generation with distributional value learning. RS-Diffuser learns a diffusion planner over future state trajectories and a separate inverse dynamics model for action recovery. Its key component is a \emph{distributional value critic}, trained with quantile regression to predict the full return distribution of candidate plans rather than only their expected value. At inference time, instead of retraining the planner for a fixed risk preference, we inject a risk-sensitive guidance signal into the diffusion denoising process, where gradients derived from tail-aware objectives such as CVaR steer generation toward desired risk profiles. As a result, a single trained model can realize risk-averse, risk-neutral, or risk-seeking behaviors by changing only the inference-time risk parameter.

Our work makes three main contributions:
\begin{itemize}
    \item[$\bullet$] We introduce a diffusion planning framework for offline RL that explicitly incorporates distributional value modeling for tail-aware decision making.
    \item[$\bullet$] We show how a distributional critic can be integrated into guided diffusion sampling, yielding flexible inference-time control over risk preferences.
    \item[$\bullet$] We demonstrate on two risk-sensitive benchmarks that RS-Diffuser outperforms existing state-of-the-art models.
\end{itemize}

\vspace{-0.2in}
\section{Related Works}
\subsubsection{Diffusion Models for Decision Making.}
The application of diffusion models for decision making has gained significant traction due to their ability to represent complex, multi-modal policy distributions~\cite{zhu2023diffusion}. Diffusion planning represents a major branch of this research, where the model generates entire sequences of future states or actions~\cite{janner2022planning}. For example, Diffuser~\cite{janner2022planning} utilizes a U-Net architecture to iteratively denoise trajectories guided by a reward function. Building on this, Decision Diffuser~\cite{ajay2022conditional} frames decision-making as a pure conditional generative task, removing the need for explicit reward gradients during sampling. Recent studies have further optimized these architectures, exploring the benefits of Transformer-based denoising networks and specialized sampling techniques~\cite{lu2025makes,chen2025toward,mejia2025integrating}. The applications of diffusion models in decision-making span diverse domains, including autonomous driving~\cite{zheng2025diffusion,zhang2026multivariate,zhang2026adaptive,jiao2026large} and sports decision-making~\cite{xu2026tacticgen}. While these methods excel at capturing the behavior distribution, they are inherently risk-neutral. They prioritize the most frequent or highest-reward paths in the training set without a mechanism to mitigate the risk of the generated plans.

\vspace{-0.1in}
\subsubsection{Risk-sensitive Reinforcement Learning.}
Risk-sensitive RL aims to learn policies that account for the uncertainty and variability of returns~\cite{noorani2025risk,ma2025dsac}. A cornerstone of this field is Distributional RL, which models the entire distribution of the random return variable instead of just its expectation~\cite{dabney2018distributional,bellemare2017distributional}. Bellemare et al.~\cite{bellemare2017distributional} introduced the categorical distribution approach, while Dabney et al.~\cite{dabney2018distributional} enhanced this with Quantile Regression DQN (QR-DQN), allowing for more flexible and accurate distribution modeling. To achieve risk-aversion, researchers often optimize metrics such as the Conditional Value-at-Risk (CVaR), which focuses on the lower tail of the return distribution~\cite{ying2022towards,ma2021conservative}. In the offline setting, risk-aversion is particularly crucial as it provides a natural defense against the epistemic uncertainty arising from limited data~\cite{chen2025diffusion}. However, traditional risk-sensitive offline RL methods are often built upon standard actor-critic architectures, which can suffer from limited policy expressivity and training instabilities in complex environments. It is worth noting that another line of research focuses on constrained learning from demonstrations, namely Inverse Constrained Reinforcement Learning (ICRL)~\cite{xu2024uncertainty,xu2024robust,liu2024comprehensive,zhao2025toward,yue2025understanding,yue2024provably}, which aims to infer latent safety constraints from expert demonstrations and incorporate them into policy optimization. Our work departs from these by embedding distributional risk-awareness directly into the generative diffusion process, leveraging the expressive power of diffusion models.

\section{Problem Formulation}\label{sec:problem_formulation}
\vspace{-0.1in}
\subsubsection{Markov Decision Process (MDP).} 
The agent optimizes the control policy under a Markov Decision Process (MDP), defined by a tuple $(\mathcal{S}, \mathcal{A}, P, r, \gamma)$, where $\mathcal{S}$ and $\mathcal{A}$ denote the state and action spaces, $P(s' \mid s,a)$ is the transition dynamics, $r(s,a)$ is the reward function, and $\gamma \in (0,1)$ is the discount factor. At each time step, the agent selects an action according to a policy $\pi(a \mid s)$, receives a reward, and transitions to the next state. The objective is to learn a policy that maximizes the expected discounted return. In this work, we consider the offline RL setting, where the policy is learned solely from a fixed dataset without further interaction with the environment.

\vspace{-0.1in}
\subsubsection{Diffusion Planner.}
Let $\mathbf{x}_0=(s_0,a_0,s_1,a_1,\dots,s_{H-1},a_{H-1},s_H)$ denote a trajectory segment of horizon $H$ sampled from an offline dataset, where $s_t\in\mathcal{S}$ and $a_t\in\mathcal{A}$. A diffusion planner~\cite{janner2022planning} models a conditional distribution over trajectories as $p(\mathbf{x}_0\mid \mathbf{c})$,
where $\mathbf{c}$ denotes the planning condition, such as the current state, a goal, or a target return. Following Denoising Diffusion Probabilistic Models (DDPM)~\cite{ho2020denoising}, the clean trajectory $\mathbf{x}_0$ is gradually corrupted into a noisy version $\mathbf{x}^k$ over diffusion steps $k=1,\dots,K$, and a denoising network is trained to reverse this process\footnote{
In this work, we use superscripts ($k \in \{0, 1, . . . , K\}$) to denote
diffusion timesteps and subscripts ($t \in \{0, 1, . . . , T\}$) to denote
trajectory timesteps}. Specifically, the learned reverse dynamics are parameterized as $p_\theta(\mathbf{x}^{k-1}\mid \mathbf{x}^k,\mathbf{c})
=
\mathcal{N}\!\big(\mu_\theta(\mathbf{x}^k,k,\mathbf{c}),\Sigma_k\big)$, where $\mu_\theta$ is a denoising neural network and $\Sigma_k$ is a predefined covariance. Training minimizes the standard denoising objective~\cite{ho2020denoising}:
\begin{equation}\label{eq:diffusion-loss}
\mathcal{L}_{\mathrm{DDPM}}(\theta)
=
\mathbb{E}_{\mathbf{x}^0,\epsilon,k}
\left[
\left\|
\epsilon-\epsilon_\theta(\mathbf{x}^k,k,\mathbf{c})
\right\|_2^2
\right],
\end{equation}
where $\mathbf{x}^k=\sqrt{\bar{\alpha}_k}\mathbf{x}^0+\sqrt{1-\bar{\alpha}_k}\epsilon$ denotes the corrupted data and $\epsilon\sim\mathcal{N}(0,\mathbf{I})$ denotes the sampled noise. At inference time, the planner starts from Gaussian noise and iteratively applies the learned reverse process via $\mu_\theta$ to recover a clean trajectory respecting $\mathbf{c}$.

\vspace{-0.1in}
\subsubsection{Guided Sampling.}
To bias generation toward a task objective $y$, we introduce a differentiable guidance model defined on the noisy trajectory, written as $p_\phi(y\mid \mathbf{x}^k,\mathbf{c},k)$. Using the standard guidance approximation~\cite{sohl2015deep,zheng2025diffusion,xu2026tacticgen}, the reverse generation step can be steered toward the desired objective as follows:
\begin{equation}\label{eq:guidance}
p_{\theta,\phi}(\mathbf{x}^{k-1}\mid \mathbf{x}^k,\mathbf{c},y)
\approx
\mathcal{N}\!\left(
\mu_\theta(\mathbf{x}^k,k,\mathbf{c})
+
w\,\Sigma_k\,g_k,\,
\Sigma_k
\right),
\end{equation}
where $w>0$ controls the guidance strength and $g_k=\nabla_{\mathbf{x}^k}\log p_\phi(y\mid \mathbf{x}^k,\mathbf{c},k)$ represents the gradient of the classifier regarding the noisy sample $\mathbf{x}^k$.
By training the diffusion model $\mu_\theta$ and the guidance model $p_\phi$ separately, the diffusion model learns a trajectory prior from offline data, while the guidance term steers the denoising process toward trajectories that better satisfy the desired objective.

\vspace{-0.1in}
\subsubsection{Distributional Value Modeling.}
In distributional RL~\cite{bellemare2017distributional,bellemare2023distributional}, the value is modeled as a random return rather than only its expectation. For a trajectory $\mathbf{x}$, let $Z(\mathbf{x})$ denote the random discounted return associated with $\mathbf{x}$. Rather than predicting only the expectation as previous offline RL methods~\cite{kumar2020conservative,wang2022diffusion,zhang2025energy}, we model the full return distribution, which is essential for risk-sensitive decision making.

Following quantile-based distributional RL~\cite{dabney2018distributional,xu2024uncertainty}, we approximate the return distribution $Z(\mathbf{x})$ by $N$ quantile values $\{z_i(\mathbf{x})\}_{i=1}^N$, associated with quantile fractions $\{\tau_i\}_{i=1}^N \subset (0,1)$. Each $\tau_i$ specifies a probability level, and the corresponding quantile value $z_i(\mathbf{x})$ approximates the return threshold below which a fraction $\tau_i$ of outcomes lie. The resulting distribution is represented as a uniform mixture of Dirac masses located at these quantile values $Z(\mathbf{x}) \approx \frac{1}{N}\sum_{i=1}^N \delta_{z_i(\mathbf{x})}$, where $\delta_{z_i(\mathbf{x})}$ denotes a Dirac delta distribution centered at $z_i(\mathbf{x})$. In practice, the quantile fractions $\{\tau_i\}$ are typically chosen uniformly in $(0,1)$.
This distributional critic enables us to compute tail aware risk measures, such as VaR~\cite{duffie1997overview} and CVaR~\cite{rockafellar2000optimization}, which can be used to guide diffusion sampling.

\section{Method}

\subsection{Overview}
Given an offline trajectory dataset $\mathcal{D}$ that many consists of state-action-reward pairs, our goal is to learn a diffusion planner that can generate high-quality future plans while allowing explicit control over risk preferences at inference time. To this end, RS-Diffuser consists of three components: 
1) a diffusion planner that generates future state trajectories conditioned on the current state (Section~\ref{sec:diffusion-planner}), 
2) an inverse dynamics model that converts planned state transitions into executable actions (Section~\ref{sec:inverse-dynamic}), and 
3) a distributional value critic that predicts the return distribution of candidate generated trajectories (Section~\ref{sec:distributional-value}). 

At test time, the distributional critic provides a differentiable risk-sensitive objective whose gradients guide the diffusion denoising process, allowing a single diffusion model to produce risk-averse, risk-neutral, or risk-seeking plans by adjusting the risk parameter (Section~\ref{sec:implementation}).

\subsection{State Trajectory Diffusion Planner}\label{sec:diffusion-planner}
While the generic formulation in Sec.~\ref{sec:problem_formulation} allows state-action trajectories, our method instantiates the diffusion variable as a \emph{future state plan} by following~\cite{ajay2022conditional}. Specifically, at decision time $t$, we define $\mathbf{x}^0=(s_{t+1},s_{t+2},\dots,s_{t+H})$, which denotes a future state sequence of planning horizon $H$, conditioned on the current state $s_t$. The planner therefore models $p_\theta(\mathbf{x}^0 \mid s_t)$.

This state-only formulation decouples long-horizon planning from low-level action generation, allowing the diffusion model to focus on predicting feasible future state evolutions while delegating action recovery to a separate inverse dynamics model. By adapting Equation~\ref{eq:diffusion-loss} to the state-only setting, the planner $\mu_\theta$ is trained with the following denoising objective:
\begin{equation}
\mathcal{L}_{\mathrm{plan}}(\theta)
=
\mathbb{E}_{\mathbf{x}^0,\epsilon,k}
\left[
\left\|
\epsilon-\epsilon_\theta(\mathbf{x}^k,k,s_t)
\right\|_2^2
\right],
\end{equation}
At inference time, the planner starts from Gaussian noise and iteratively denoises to produce a future state plan $\hat{\mathbf{x}}^0=(\hat{s}_{t+1},\dots,\hat{s}_{t+H})$.

\subsection{Inverse Dynamics Action Decoder}\label{sec:inverse-dynamic}
Because the planner operates in state space, we use a separate inverse dynamics model to recover executable actions from planned state transitions. Specifically, we learn a deterministic decoder $f_\psi(s_t,s_{t+1})$ for continuous control. In practice, given the current state $s_t$ and the predicted next state $\hat{s}_{t+1}$ from the diffusion planner, the executed action is obtained as $\hat{a}_t = f_\psi(s_t,\hat{s}_{t+1})$. The inverse dynamics model is trained on offline transitions using supervised regression~\cite{ajay2022conditional}:
\begin{equation}\label{eq:inv}
\mathcal{L}_{\mathrm{inv}}(\psi)
=
\mathbb{E}_{(s_t,a_t,s_{t+1})\sim\mathcal{D}}
\left[
\|a_t-f_\psi(s_t,s_{t+1})\|_2^2
\right].
\end{equation}
This design separates \emph{what} future states should be reached from \emph{how} to realize them, which is particularly suitable for diffusion planning over long horizons.

\subsection{Monte Carlo Distributional Value Critic}\label{sec:distributional-value}

The core of RS-Diffuser is a distributional value critic that estimates the return-to-go distribution of candidate plans. Unlike standard risk-neutral value guidance, which predicts only the expected return, our critic models the full return distribution and therefore supports tail-aware risk measures.

For each trajectory segment starting at time step $t$, we compute the Monte Carlo discounted return-to-go as:
\begin{equation}\label{eq:return}
R_t=\sum_{\ell=0}^{T_e-t-1}\gamma^\ell r_{t+\ell},
\end{equation}
where $\gamma\in(0,1]$ is the discount factor and $T_e$ denotes the terminal index.

To make the critic compatible with denoising-time guidance, we condition it on the noisy plan $\mathbf{x}^k$, the current state $s_t$, and the diffusion step $k$. The distributional critic then predicts $N$ quantiles $\mathbf{z}_\phi(\mathbf{x}^k,s_t,k)
=
\left\{
z_i^\phi(\mathbf{x}^k,s_t,k)
\right\}_{i=1}^N$,
corresponding to quantile fractions $\{{\tau}_i\}_{i=1}^N\subset(0,1)$. Importantly, we adopt Monte Carlo supervision rather than Bellman bootstrapping, following the value-learning paradigm in~\cite{lu2025makes}. Specifically, we train the critic by quantile regression using the Monte Carlo return sample $R_t$:
\begin{equation}\label{eq:value-critic}
\mathcal{L}_{\mathrm{dist}}(\phi)
=
\mathbb{E}_{(\mathbf{x}^0,s_t,R_t)\sim\mathcal{D},\,\epsilon,\,k}
\left[
\frac{1}{N}\sum_{i=1}^N
\rho_{{\tau}_i}^{\kappa}
\!\left(
R_t-z_i^\phi(\mathbf{x}^k,s_t,k)
\right)
\right],
\end{equation}
where $\mathcal{D}$ denotes the offline dataset and $\rho_{\tau}^{\kappa}(\cdot)$ is the quantile Huber loss~\cite{dabney2018distributional}:
\begin{equation}
\rho_{\tau}^{\kappa}(u)
=
\left|\tau-\mathbb{I}(u<0)\right|
L_\kappa(u), \quad L_\kappa(u)=
\begin{cases}
\frac{1}{2}u^2/\kappa, & |u|\le \kappa,\\[2mm]
|u|-\frac{1}{2}\kappa, & \text{otherwise.}
\end{cases}
\end{equation}

By learning the full return distribution instead of only its mean, the critic provides a natural basis for risk-sensitive guidance during inference.

\subsection{Practical Implementation of RS-Diffuser}\label{sec:implementation}
\subsubsection{Risk Sensitive Objective from Quantiles.}
Once the return distribution is available, different risk preferences can be expressed through different functionals of the predicted quantiles. For a risk-averse objective, we use the lower-tail CVaR, approximated by averaging quantiles up to level $\alpha$:
\begin{equation}
\rho_{\alpha}^{\mathrm{low}}(\mathbf{x}^k,s_t,k)
=
\mathrm{CVaR}_{\alpha}(\mathbf{x}^k,s_t,k)
\approx
\frac{1}{|\mathcal{I}_{\alpha}|}
\sum_{i\in\mathcal{I}_{\alpha}}
z_i^\phi(\mathbf{x}^k,s_t,k),
\end{equation}
where $\mathcal{I}_{\alpha}=\{i:\tau_i\le \alpha\}$. Maximizing this objective encourages plans with favorable worst-case outcomes. For risk-seeking behavior we use an upper-tail objective:
\begin{equation}
\rho_{\alpha}^{\mathrm{up}}(\mathbf{x}_k,s_t,k)
=
\frac{1}{|\mathcal{J}_{\alpha}|}
\sum_{i\in\mathcal{J}_{\alpha}}
z_i^\phi(\mathbf{x}^k,s_t,k),
\qquad
\mathcal{J}_{\alpha}=\{i:\tau_i\ge 1-\alpha\}.
\end{equation}

\subsubsection{Risk Guided Diffusion Sampling}
We now instantiate the generic guidance mechanism in Sec.~\ref{sec:problem_formulation} with the distributional value critic. Instead of using a conventional classifier, we treat the risk functional $\rho_\alpha(\mathbf{x}^k,s_t,k)$ as a differentiable guidance objective defined on the noisy plan. Its gradient with respect to the current denoising state is:
\begin{equation}\label{eq:gradient}
g_k = \nabla_{\mathbf{x}^k}\rho_\alpha(\mathbf{x}^k,s_t,k),
\end{equation}
where $\rho_\alpha$ denotes the used risk measure on the distributional critic. The guided reverse process can then be calculated by Eq.~\ref{eq:guidance}. Intuitively, the critic-derived gradient biases denoising toward trajectories that match the desired risk profile.
\vspace{-0.1in}

\begin{algorithm}[htbp]
\caption{RS-Diffuser}
\label{alg:rsdiffuser}
\small
\begin{algorithmic}[1]
\REQUIRE Offline dataset $\mathcal{D}$, planning horizon $H$, diffusion steps $K$, quantile numbers $N$, risk objective $\rho_\alpha$

\STATE Calculate accumulated discounted returns for every step $t$ via Equation~\ref{eq:return}.

\STATE \textbf{Offline training}
\FOR{each training iteration}
    \STATE Sample $(s_t,\mathbf{x}^0,R_t)$ from $\mathcal{D}$, where $\mathbf{x}^0=(s_{t+1},\dots,s_{t+H})$
    \STATE Update diffusion planner $\epsilon_\theta$ with denoising loss $\mathcal{L}_{\mathrm{plan}}$ via Equation~\ref{eq:diffusion-loss}
    \STATE Update inverse dynamics model $f_\psi$ with loss $\mathcal{L}_{\mathrm{inv}}$ via Equation~\ref{eq:inv}
    \STATE Update distributional critic $z_\phi$ with quantile loss $\mathcal{L}_{\mathrm{dist}}$ via Equation~\ref{eq:value-critic}
\ENDFOR

\STATE \textbf{Risk-guided planning at inference}
\FOR{each decision step}
    \STATE Observe current state $s_t$ and initialize $\mathbf{x}^K \sim \mathcal{N}(0,\mathbf{I})$
    \FOR{$k=K,K-1,\dots,1$}
        \STATE Compute risk guidance $g_k=\nabla_{\mathbf{x}^k}\rho_\alpha(\mathbf{x}^k,s_t,k)$ using $z_\phi$ via Equation~\ref{eq:gradient}
        \STATE Perform guided reverse diffusion to obtain $\mathbf{x}^{k-1}$
    \ENDFOR
    \STATE Obtain planned states $\hat{\mathbf{x}}^0=(\hat{s}_{t+1},\dots,\hat{s}_{t+H})$
    \STATE Decode action $\hat{a}_t=f_\psi(s_t,\hat{s}_{t+1})$ and execute it
\ENDFOR
\end{algorithmic}
\end{algorithm}
\vspace{-0.2in}

\subsubsection{Remarks.} It is worth noting that this design is a key advantage of RS-Diffuser: the planner is trained only once, while risk preference is entirely controlled at inference time through the choice of $\rho_\alpha$ applied to the distributional value critic. As a result, a single model can seamlessly switch between conservative, neutral, and aggressive behaviors according to different requirements without retraining. Algorithm~\ref{alg:rsdiffuser} summarizes the overall pipeline of RS-Diffuser, including offline training of the planner, inverse dynamics model, and distributional critic, as well as risk-guided planning at inference time.

\section{Experiments}
In this section, we empirically evaluate the proposed RS-Diffuser against state-of-the-art methods on risk-sensitive benchmarks, including risk-sensitive D4RL (Section~\ref{sec:d4rl}) and risky robot navigation (Section~\ref{sec:nav}). To further assess robustness, we conduct ablation studies on key hyperparameters (Section~\ref{sec:ablation}).

\vspace{-0.1in}
\subsection{Experimental Setup}

\subsubsection{Model Configurations} We implement our method using CleanDiffuser~\cite{dong2024cleandiffuser}. The planner uses a 2-block Transformer (hidden size 256), and the inverse dynamics model is an MLP (hidden size 256). We adopt DDIM~\cite{song2020denoising} with 20 steps for planning and DDPM~\cite{ho2020denoising} with 10 steps for inverse dynamics. Models are trained for 1M steps with batch size 256. The planning horizon is 32 with discount factor 0.99. We use Adam optimizer with a learning rate 3e-4.

\vspace{-0.1in}
\subsubsection{Baseline Methods} We compare with risk neutral baselines, Conservative Q Learning (CQL~\cite{kumar2020conservative}) and Diffusion QL~\cite{wang2022diffusion}, as well as risk sensitive methods:
1) OWCPG~\cite{tang2019worst}, optimizing worst case (e.g., CVaR) returns;
2) ORAAC~\cite{urpi2021risk}, a distributional critic with an imitation based policy;
3) CODAC~\cite{ma2021conservative}, conservative offline RL with a distributional critic;
4) UDAC~\cite{chen2025diffusion}, an uncertainty aware actor critic with a diffusion policy.

\vspace{-0.1in}
\subsubsection{Evaluation Metric} Each method is evaluated over 100 test episodes. We report episodic return (cumulative reward per episode) as the main metric, including both the mean and CVaR$_{0.1}$ (average of the worst 10 episodes). For risky navigation tasks, we also report episodic violations, defined as the total number of time steps spent in risky regions. Results are averaged over five random seeds and reported as mean $\pm$ standard deviation.

\subsection{Risk-sensitive D4RL}\label{sec:d4rl}

\subsubsection{Task Description} The risk-sensitive D4RL benchmark~\cite{urpi2021risk} introduces stochastic penalties in MuJoCo to model safety-critical failures. Rewards are reduced by large penalties when agents enter undesirable states, such as exceeding velocity thresholds in Half-Cheetah or violating stability constraints (e.g., large pitch angles) in Walker2D and Hopper. This creates rare but severe negative outcomes, requiring policies to balance performance and safety. The maximum episode length is 500 steps.

\begin{table*}[htbp]
\caption{Episodic return on risk-sensitive D4RL tasks. Higher is better for both CVaR$_{0.1}$ and Mean metrics.}
\label{tab:d4rl}
\centering
\small
\setlength{\tabcolsep}{5pt}
\resizebox{\textwidth}{!}{
\begin{tabular}{c l | cc | cc | cc}
\toprule
& \multirow{2}{*}{Algorithm}
& \multicolumn{2}{c|}{Medium}
& \multicolumn{2}{c|}{Expert}
& \multicolumn{2}{c}{Mixed} \\
& 
& CVaR$_{0.1}\uparrow$ & Mean$\uparrow$
& CVaR$_{0.1}\uparrow$ & Mean$\uparrow$
& CVaR$_{0.1}\uparrow$ & Mean$\uparrow$ \\
\midrule

\multirow{7}{*}{\rotatebox[origin=c]{90}{\bf Half-Cheetah}}
& CQL          & -15$\pm$17   & 33$\pm$36    & -207$\pm$47   & -75$\pm$23    & 12$\pm$24    & 214$\pm$52 \\
& Diffusion-QL & 84$\pm$32    & 329$\pm$20   & 542$\pm$102   & 728$\pm$104   & 143$\pm$77   & 199$\pm$43 \\
& OWCPG        & 76$\pm$14    & 316$\pm$23   & 248$\pm$232   & 905$\pm$107   & 164$\pm$76   & 217$\pm$33 \\
& ORAAC        & 214$\pm$36   & 331$\pm$30   & 595$\pm$191   & 1180$\pm$78   & 119$\pm$27   & 307$\pm$6 \\
& CODAC        & -41$\pm$17   & 338$\pm$25   & 687$\pm$152   & 1255$\pm$101  & 238$\pm$59   & 396$\pm$56 \\
& UDAC         & 276$\pm$29   & 417$\pm$18   & 732$\pm$104   & 1352$\pm$100  & 275$\pm$50   & 462$\pm$48 \\
& RS-Diffuser  & \textbf{292$\pm$21}   & \textbf{431$\pm$29}   & \textbf{793$\pm$125}   & \textbf{1409$\pm$113}  & \textbf{301$\pm$62}   & \textbf{539$\pm$56} \\
\midrule

\multirow{7}{*}{\rotatebox[origin=c]{90}{\bf Walker-2D}}
& CQL          & 1244$\pm$128 & 1524$\pm$99  & 1301$\pm$78   & 2018$\pm$65   & 74$\pm$77    & -64$\pm$78 \\
& Diffusion-QL & 31$\pm$10    & 273$\pm$20   & 621$\pm$49    & 1202$\pm$63   & 302$\pm$154  & 102$\pm$88 \\
& OWCPG        & -15$\pm$41   & 283$\pm$37   & 362$\pm$33    & 1372$\pm$160  & 201$\pm$33   & -77$\pm$54 \\
& ORAAC        & 663$\pm$124  & 1134$\pm$20  & 1172$\pm$71   & 2006$\pm$56   & 222$\pm$37   & -70$\pm$76 \\
& CODAC        & 1159$\pm$57  & 1537$\pm$78  & 1298$\pm$98   & 2102$\pm$102  & 450$\pm$193  & 261$\pm$231 \\
& UDAC         & 1368$\pm$89  & \textbf{1602$\pm$85}  & 1398$\pm$67   & 2198$\pm$78   & 498$\pm$91   & 299$\pm$194 \\
& RS-Diffuser  & \textbf{1493$\pm$107} & 1588$\pm$74   & \textbf{1406$\pm$83}   & \textbf{2204$\pm$110}  & \textbf{531$\pm$84}   & \textbf{377$\pm$146} \\
\midrule

\multirow{7}{*}{\rotatebox[origin=c]{90}{\bf Hopper}}
& CQL          & 1344$\pm$40  & 1524$\pm$20  & 1289$\pm$30   & 1402$\pm$30   & 189$\pm$63   & -21$\pm$62 \\
& Diffusion-QL & 981$\pm$28   & 1001$\pm$11  & 406$\pm$31    & 583$\pm$18    & 1021$\pm$102 & 1010$\pm$59 \\
& OWCPG        & -87$\pm$25   & 69$\pm$8     & 720$\pm$34    & 898$\pm$12    & 372$\pm$100  & 442$\pm$201 \\
& ORAAC        & 1416$\pm$28  & 1482$\pm$4   & 980$\pm$28    & 1385$\pm$33   & 876$\pm$87   & 525$\pm$323 \\
& CODAC        & 976$\pm$30   & 1014$\pm$16  & 990$\pm$19    & 1398$\pm$29   & 1551$\pm$33  & 1450$\pm$101 \\
& UDAC         & 1502$\pm$19  & 1602$\pm$18  & 1302$\pm$28   & 1502$\pm$27   & 1620$\pm$42  & 1523$\pm$90 \\
& RS-Diffuser  & \textbf{1590$\pm$31}   & \textbf{1732$\pm$23}   & \textbf{1475$\pm$39}   & \textbf{1623$\pm$39}  & \textbf{1702$\pm$69}   & \textbf{1609$\pm$134} \\
\bottomrule
\end{tabular}
}
\vspace{-0.2in}
\end{table*}

\subsubsection{Results Analysis} Table~\ref{tab:d4rl} shows the results. We find that RS-Diffuser achieves the best overall performance on the risk-sensitive D4RL benchmark, outperforming both risk-neutral and risk-sensitive baselines across most settings. This indicates not only stronger average performance but also improved robustness under worst-case scenarios. In contrast, risk-neutral methods such as CQL perform poorly, as optimizing only the expected return fails to capture tail risks in safety-critical environments. Diffusion-QL achieves moderately better performance, benefiting from the expressive generative capacity of diffusion models, but still lacks explicit risk awareness. Risk-sensitive baselines such as UDAC demonstrate relatively strong performance. However, they are typically trained to optimize a fixed risk objective, limiting their adaptability across different scenarios. In contrast, RS-Diffuser enables flexible control over risk preferences at inference time, allowing a single model to adapt to different risk measures and consistently achieve strong performance. Overall, these results demonstrate that guiding diffusion planning with a distributional critic is an effective approach for improving tail-aware decision making in offline RL.

\subsection{Risky Robot Navigation}\label{sec:nav} 
\subsubsection{Task Description} Risky robot navigation benchmark was introduced in~\cite{ma2021conservative}. It was evaluated on two challenging navigation tasks: Risky PointMass and Risky Ant. In both environments, the agent starts from a randomly initialized state and aims to reach a target location efficiently.

\begin{wrapfigure}{r}{0.175\textwidth}
\centering
\vspace{-0.35in}
\includegraphics[width=0.175\textwidth]{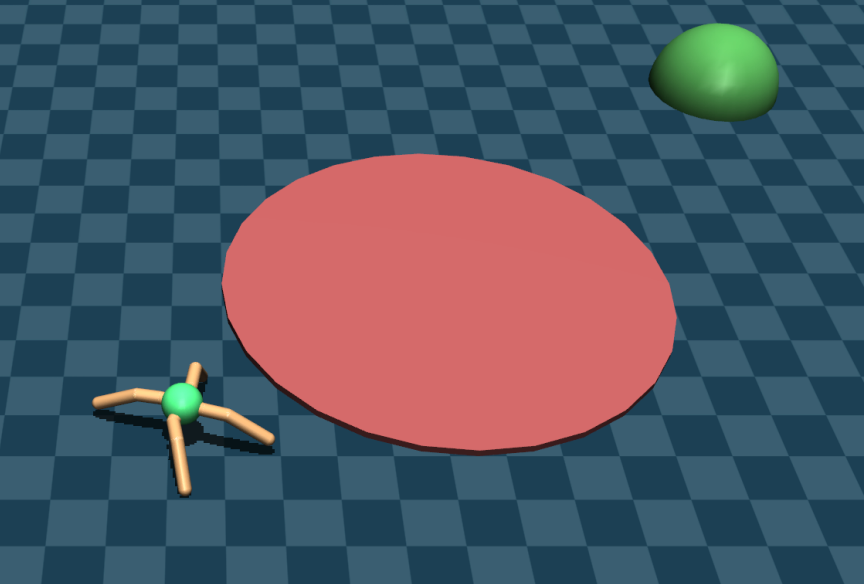}
\vspace{-0.35in}
\caption{Risky Ant.}
\label{fig:riskyant}
\vspace{-0.2in}
\end{wrapfigure}
However, the direct path to the goal passes through a hazardous region that imposes significant penalties. While risk-neutral policies tend to take the shorter but hazardous route, risk-sensitive policies are expected to avoid the hazardous area and instead favor safer trajectories. Figure~\ref{fig:riskyant} shows an illustrative example of Risky Ant.

\begin{table*}[htbp]
\caption{Episodic return and violations on risky robot navigation tasks. For return metrics (Mean, Median, and CVaR$_{0.1}$), higher is better. For violations, lower is better.}
\label{tab:risky_robot_navigation}
\vspace{-0.1in}
\centering
\small
\setlength{\tabcolsep}{6pt}
\resizebox{\textwidth}{!}{
\begin{tabular}{l|cccc|cccc}
\toprule
\multirow{2}{*}{Algorithm}
& \multicolumn{4}{c|}{Risky PointMass}
& \multicolumn{4}{c}{Risky Ant} \\
& Mean $\uparrow$ & Median $\uparrow$ & CVaR$_{0.1}$ $\uparrow$ & Violations $\downarrow$
& Mean $\uparrow$ & Median $\uparrow$ & CVaR$_{0.1}$ $\uparrow$ & Violations $\downarrow$ \\
\midrule
CQL          & -7.5$\pm$2.0   & -4.4$\pm$1.6   & -43.4$\pm$2.8   & 93.4$\pm$10.5
             & -967.8$\pm$78.2 & -858.5$\pm$87.2 & -1887.3$\pm$122.4 & 1854.3$\pm$130.2 \\
Diffusion-QL & -13.5$\pm$2.2  & -5.7$\pm$1.3   & -69.2$\pm$4.6   & 140.1$\pm$10.5
             & -892.5$\pm$80.3 & -766.5$\pm$66.7 & -1884.4$\pm$133.2 & 1902.4$\pm$127.9 \\
OWCPG       & -12.4$\pm$2.5  & -5.1$\pm$1.7   & -67.4$\pm$4.0   & 123.5$\pm$10.6
             & -819.2$\pm$77.4 & -791.5$\pm$80.2 & -1424.5$\pm$98.2  & 1455.3$\pm$104.5 \\
ORAAC        & -10.7$\pm$1.6  & -4.6$\pm$1.3   & -64.1$\pm$2.6   & 138.7$\pm$20.2
             & -788.1$\pm$98.2 & -795.3$\pm$86.2 & -1247.2$\pm$105.7 & 1196.1$\pm$99.6 \\
CODAC        & -6.1$\pm$1.8   & -4.9$\pm$1.2   & -14.7$\pm$2.4   & 0$\pm$0
             & -456.3$\pm$53.2 & -433.4$\pm$47.2 & -686.6$\pm$87.2  & 347.8$\pm$42.1 \\
UDAC         & -5.8$\pm$1.4   & -4.2$\pm$1.3   & -12.4$\pm$2.0   & 0$\pm$0
             & -392.5$\pm$33.2 & -399.5$\pm$40.5 & -598.5$\pm$50.2  & 291.5$\pm$20.8 \\
RS-Diffuser  & \textbf{-4.7$\pm$1.1}   & \textbf{-3.6$\pm$1.0}   & \textbf{-9.9$\pm$2.5}   & \textbf{0$\pm$0}
             & \textbf{-365.1$\pm$33.2} & \textbf{-371.5$\pm$43.7} & \textbf{-574.3$\pm$53.1}  & \textbf{269.0$\pm$17.5} \\
\bottomrule
\end{tabular}
}
\vspace{-0.1in}
\end{table*}

\vspace{-0.1in}
\subsubsection{Results Analysis} Table~\ref{tab:risky_robot_navigation} shows the results. In the Risky PointMass environment, CODAC, UDAC, and the proposed RS-Diffuser all achieve zero violations while maintaining high returns, reflecting the relative simplicity of controlling a point mass system. Despite the absence of violations across these methods, RS-Diffuser consistently achieves the highest performance in terms of Mean, Median, and CVaR$_{0.1}$, indicating superior overall return and stronger tail performance. In the more challenging Risky Ant environment, RS-Diffuser attains the lowest number of violations while simultaneously achieving the highest returns. This demonstrates its ability to effectively balance safety and performance. The results highlight the advantage of combining a distributional value critic for risk-aware evaluation with the expressive power of diffusion models for modeling complex transitions.

\vspace{-0.1in}
\subsection{Ablation Study on Hyperparameters}\label{sec:ablation}

\subsubsection{Task Description} To further evaluate the robustness of the proposed method, we conduct an ablation study on the Half-Cheetah environment by varying the guidance measure $\rho_\alpha$. Specifically, we consider both VaR and CVaR as risk measures and systematically vary the parameter $\alpha$ to model different risk preferences. Smaller values of $\alpha$ correspond to more risk-averse behavior, while larger values encourage more risk-seeking behavior.

\begin{table}[htbp]
\vspace{-0.1in}
\caption{Ablation on risk-sensitive Half-Cheetah with varying risk measures $\rho_\alpha$ at inference time. CVaR$_{0.1}$ is used as the default in the main experiments.}
\label{tab:ablation}
\vspace{-0.1in}
\centering
\small
\setlength{\tabcolsep}{6pt}
\resizebox{\textwidth}{!}{
\begin{tabular}{l|cc|cc|cc}
\toprule
\multirow{2}{*}{$\rho_\alpha$}
& \multicolumn{2}{c|}{Medium}
& \multicolumn{2}{c|}{Expert}
& \multicolumn{2}{c}{Mixed} \\
& CVaR$_{0.1}\uparrow$ & Mean$\uparrow$
& CVaR$_{0.1}\uparrow$ & Mean$\uparrow$
& CVaR$_{0.1}\uparrow$ & Mean$\uparrow$ \\
\midrule
CVaR$_{0.05}$          & \textbf{305$\pm$19} & 408$\pm$24 & \textbf{826$\pm$118} & 1361$\pm$105 & \textbf{319$\pm$48} & 512$\pm$53 \\
CVaR$_{0.1}$ (default) & 292$\pm$21          & 431$\pm$29 & 793$\pm$125          & 1409$\pm$113 & 301$\pm$62          & 539$\pm$56 \\
CVaR$_{0.2}$           & 271$\pm$24          & 452$\pm$31 & 742$\pm$132          & 1463$\pm$120 & 278$\pm$67          & 563$\pm$60 \\
VaR$_{0.05}$           & 286$\pm$23          & 418$\pm$27 & 775$\pm$129          & 1388$\pm$111 & 292$\pm$54          & 526$\pm$55 \\
VaR$_{0.1}$            & 279$\pm$25          & 439$\pm$30 & 759$\pm$131          & 1432$\pm$118 & 284$\pm$58          & 548$\pm$57 \\
VaR$_{0.2}$            & 252$\pm$29          & \textbf{463$\pm$35} & 711$\pm$140          & \textbf{1491$\pm$126} & 259$\pm$61          & \textbf{577$\pm$63} \\
\bottomrule
\end{tabular}
}
\vspace{-0.2in}
\end{table}

\subsubsection{Results Analysis} Table~\ref{tab:ablation} studies the effect of different inference-time risk measures on Half-Cheetah. As expected, smaller $\alpha$ leads to more risk-averse behavior, yielding higher CVaR$_{0.1}$ but lower mean return, while larger $\alpha$ encourages more risk-seeking behavior, improving the mean at the cost of weaker tail performance. Comparing VaR and CVaR, we observe that CVaR-based guidance generally provides stronger worst-case performance, whereas VaR-based guidance is slightly more aggressive and can achieve higher mean returns for larger $\alpha$. Among all settings, CVaR$_{0.1}$ offers the best overall trade-off between average return and robustness, which justifies its use as the default choice.

\vspace{-0.1in}
\section{Conclusion}
We presented \textbf{RS-Diffuser}, a risk-sensitive offline diffusion planning framework that integrates diffusion-based trajectory generation with distributional value critics. By guiding the denoising process with gradients derived from tail-aware risk measures such as CVaR, our method enables a single trained model to flexibly realize risk-averse, risk-neutral, and risk-seeking behaviors at inference time. Experiments on risk-sensitive D4RL and risky robot navigation benchmarks demonstrate that RS-Diffuser achieves strong performance, improved robustness, and fewer safety violations compared to prior methods. These results highlight the effectiveness of combining diffusion planning with distributional value guidance for risk-sensitive decision making. A promising direction for future work is to extend RS-Diffuser to more challenging settings, including higher-dimensional real-world robotics tasks, richer uncertainty-aware critics, and multi-objective safety constraints. We hope this work provides a useful step toward flexible and reliable risk-sensitive decision-making with diffusion models.

\bibliographystyle{splncs04}
\bibliography{references}

\end{document}